\documentclass[journal]{IEEEtran}
\usepackage{cite}
\usepackage{graphicx}
\usepackage{bm}
\usepackage{epsfig}
\usepackage{amsmath}
\usepackage{array}
\usepackage{multirow}
\usepackage{color}
\usepackage{booktabs}
\pdfoutput=1
\usepackage{balance}
\usepackage{amsfonts}
\usepackage{amssymb,bm} % define this before the line numbering.

\usepackage{algpseudocode}

\usepackage[ruled,linesnumbered]{algorithm2e}
\usepackage{listings}

\newcolumntype{C}[1]{>{\centering\arraybackslash}m{#1}}
\newcolumntype{R}[1]{>{\raggedleft\arraybackslash}m{#1}}
\newcolumntype{P}[1]{>{\raggedright\arraybackslash}p{#1}}
\newcolumntype{M}[1]{>{\centering\arraybackslash}m{#1}}

\usepackage{url}            % simple URL typesetting
\ifCLASSINFOpdf
\else
\fi

\begin{document}
%
% paper title
% Titles are generally capitalized except for words such as a, an, and, as,
% at, but, by, for, in, nor, of, on, or, the, to and up, which are usually
% not capitalized unless they are the first or last word of the title.
% Linebreaks \\ can be used within to get better formatting as desired.
% Do not put math or special symbols in the title.
\title{{Style Interleaved Learning for Generalizable Person Re-identification}}

\author{Wentao Tan,
        Changxing Ding,
       Pengfei Wang,
       Mingming Gong,
       Kui Jia

\thanks{
This work was supported by the National Natural Science Foundation of China under Grant 62076101, Guangdong Basic and Applied Basic Research Foundation under Grant 2023A1515010007, the Guangdong Provincial Key Laboratory of Human Digital Twin under Grant 2022B1212010004, and the Program for Guangdong Introducing Innovative and Entrepreneurial Teams under Grant 2017ZT07X183.
(Corresponding author: Changxing Ding.)
}

\thanks{Wentao Tan is with the School of Future Technology, South China University of Technology, 381 Wushan Road, Tianhe District, Guangzhou 510000, P.R. China (e-mail: ftwentaotan@mail.scut.edu.cn).}

\thanks{Changxing Ding and Kui Jia are with the School of Electronic and Information Engineering, South China University of Technology, 381 Wushan Road, TianheDistrict, Guangzhou 510000. PR. China, and Changxing Ding is also with the Pazhou Lab, Guangzhou 510330, China (e-mail: chxding@scut.edu.cn; kuijia@scut.edu.cn).}

\thanks{Pengfei Wang is with the Department of Computing, The Hong Kong Polytechnic University, Hung Hom, HongKong (e-mail:pengfei.wang@connect.polyu.hk).}

\thanks{Mingming Gong is with the School of Mathematics and Statistics, The University
of Melbourne, Melbourne, VIC 3010, Australia (e-mail:
mingming.gong@unimelb.edu.au).}
}

% make the title area
\maketitle

\begin{abstract}
Domain generalization (DG) for person re-identification (ReID) is a challenging problem, as access to target domain data is not permitted during the training process. Most existing DG ReID methods update the feature extractor and classifier parameters based on the same features. This common practice causes the model to overfit to existing feature styles in the source domain, resulting in sub-optimal generalization ability on target domains. To solve this problem, we propose a novel style interleaved learning (IL) framework. Unlike conventional learning strategies, IL incorporates two forward propagations and one backward propagation for each iteration. We employ the features of interleaved styles to update the feature extractor and classifiers using different forward propagations, which helps to prevent the model from overfitting to certain domain styles. 
To generate interleaved feature styles, we further propose a new feature stylization approach. It produces a wide range of meaningful styles that are both different and independent from the original styles in the source domain, which caters to the IL methodology.
Extensive experimental results show that our model not only consistently outperforms state-of-the-art methods on large-scale benchmarks for DG ReID, but also has clear advantages in computational efficiency. The code is available at \textcolor{blue}{\url{https://github.com/WentaoTan/Interleaved-Learning}}.

\end{abstract}

% Note that keywords are not normally used for peerreview papers.
\begin{IEEEkeywords}
Person Re-identification, Domain Generalization, Interleaved Learning.
\end{IEEEkeywords}

\IEEEpeerreviewmaketitle

\section{Introduction}
\begin{figure}[htp]
\begin{center}
\setlength{\abovecaptionskip}{0pt}
\includegraphics[width=\linewidth]{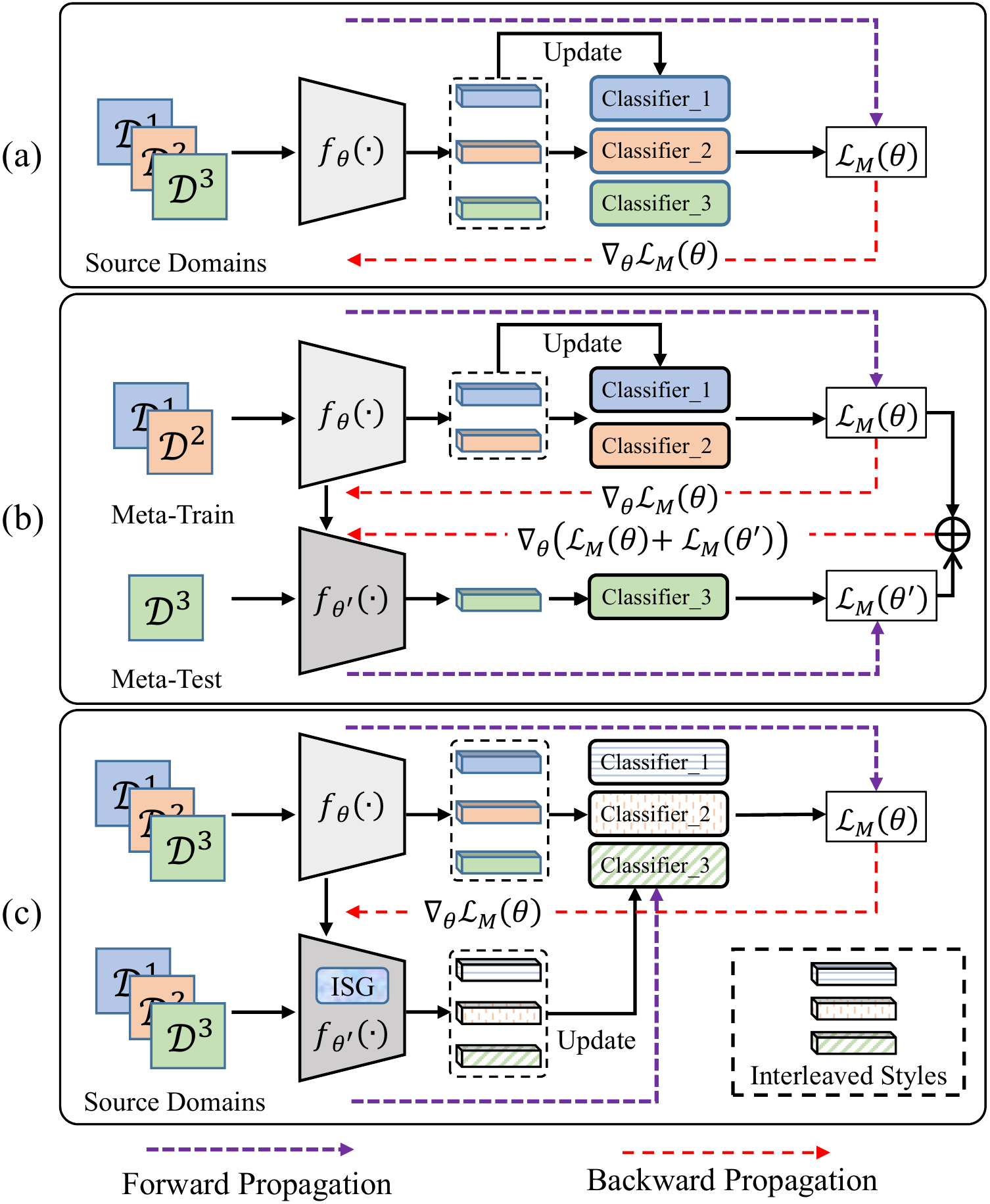}
\end{center}
\vspace{-1em}
\caption{
Differences between IL and existing learning schemes. We take the memory bank-based classifier as an example for illustrative purposes.
(a) Conventional methods update the feature extractor and multiple classifiers based on the same features. (b) Meta-learning approaches divide multiple source domains into meta-train and meta-test domains but still utilize the same features to update the feature extractor and classifiers. (c) IL updates the feature extractor and classifiers based on the features of the different styles. ISG is a powerful feature stylization approach, the details of which will be introduced in Section \ref{FS}. Best viewed in color.
}

\label{fig:example}
\vspace{-1em}
\end{figure}

\IEEEPARstart {T}{he} goal of person re-identification (ReID) is to identify images of the same person across multiple cameras. Due to its wide range of applications, which include seeking persons of interest (\textit{e.g.}, lost children), ReID research has undergone explosive growth in recent years \cite{wang2022nformer,zhu2022dual,zhuang2020rethinking,sun2018beyond,18_xu2018attention,luo2019bag,yan2021beyond,zhao2020deep,zhang2022implicit}. Most existing approaches can achieve remarkable performance when the training and testing data are drawn from the same domain. However, when these ReID models are applied to other domains (such as person images captured by a new camera system), they often suffer from clear performance drops due to the presence of domain gaps.
To alleviate these problems, domain generalization (DG) for person ReID has recently emerged as an important research topic \cite{qi2022novel,song2019generalizable,choi2021meta,dai2021generalizable,zhao2021learning,xu2022meta,zhang2022acl,jiao2022dtin}. DG ReID methods utilize labeled data from source domains to learn a generalizable model for unseen target domains. Compared with unsupervised domain adaptation (UDA) \cite{yang2020part,zhang2022implicit,dai2021idm,ge2019mutual,bai2021unsupervised,wang2022uncertainty}, the DG task is more challenging, as it cannot access any images in the target domain for model training purposes. Moreover, unlike the traditional DG setting \cite{wang2023sharpness,wan2022meta,pandey2021generalization,xu2021fourier,lv2022causality,kang2022style,lee2022wildnet}, which assumes that both domains share the same classes, DG ReID is a more challenging open-set problem, in that there is no identity overlap between any two domains.

Most DG ReID methods \cite{song2019generalizable,jin2020style,xu2022meta,zhang2022acl,jiao2022dtin} adopt a single shared feature extractor and assign a separate classifier to each source domain. As shown in Fig.~\ref{fig:example}(a), the features of each domain extracted by the feature extractor are also used to update the parameters of the corresponding classifier. We contend that this common practice leads to the model displaying sub-optimal generalization ability on unseen domains, since both the feature extractor (``player'') and classifiers (``referees'') are biased towards the same styles. We further provide an example in Section \ref{related_work} that vividly illustrates how this common practice affects the model's generalization ability. Some ReID methods adopt meta-learning \cite{li2018learning,zhao2021learning,choi2021meta,zhang2022acl}, which involves dividing multiple source domains into meta-train and meta-test domains to simulate real train-test domain shifts; however, the above issue persists under these circumstances, as illustrated in Fig. ~\ref{fig:example}(b).
During the meta-learning training process, the classifier for each domain is still updated based on the same features as those used for loss computation. 

To overcome the above limitations, we introduce a novel style interleaved learning (IL) framework for DG. As shown in Fig.~\ref{fig:example}(c), this framework adopts the features of interleaved styles to update the parameters of the feature extractor and classifiers. In more detail, there are two forward propagations and one backward propagation for each iteration. In the two forward propagations, we use the features of original styles for loss computation, then adopt the features of  synthesized styles to update the memory-based classifiers, which creates an artificial domain shift between the feature extractor and classifiers. This results in style-robust gradients in the backward propagation, thereby promoting the generalization ability of the feature extractor. Moreover, as revealed in the experimentation section, the second forward propagation is quite efficient, introducing only a very small computational cost.

In our IL framework, a stylization approach is required to produce  interleaved styles. The term ``interleaved styles" denotes that two groups of styles are independent from each other. Existing feature stylization methods (\textit{e.g.}, MixStyle \cite{zhou2021domain} and DSU \cite{li2022uncertainty}) are powerful for data augmentation. However, the styles synthesized by these methods remain closely related to those of the training samples, which deviates from the spirit of IL. For example, MixStyle \cite{zhou2021domain} produces new feature styles by mixing the styles of two samples in a linear manner, while DSU \cite{li2022uncertainty} modifies the original feature styles by adding Gaussian noise to the feature statistics of each image.
To address this issue, we propose a simple yet effective stylization method called Interleaved Style Generator (ISG). ISG estimates the range of meaningful styles based on the distribution of feature styles of the training data. It then performs uniform sampling within the estimated intervals, which ensures that each style obtained is both meaningful and independent from the original styles in the training data. 
%It is important to note that both of these benefits, namely variety and independence, can help IL to improve the generalization of the network. Specifically, diversity enriches the artificial domain shift between feature extractors and classifiers, resulting in style-robust gradients; independence prevents the network from learning misleading semantic-stylistic relations and boosts feature extractor discriminative power. 
In the experimentation section, we demonstrate that ISG outperforms existing works \cite{nuriel2021permuted,zhou2021domain,li2022uncertainty} by large margins under the IL framework.

In summary, the main contributions of this paper are three-folds: 

	(1)	We propose a novel IL framework for domain generalization, which is highly computationally efficient and can be readily applied to various model architectures.
	
	(2)	We propose a new feature stylization approach, the Interleaved Style Generator. Compared with existing stylization techniques, ISG produces feature styles that are not only more diverse but also compatible with our IL framework.

	(3)	We perform extensive experiments on multiple DG ReID benchmarks, which show that our approach consistently outperforms state-of-the-art methods by significant margins.

The remainder of this paper is structured as follows. In Section \ref{related_work}, related works on DG for person ReID and IL are briefly reviewed. Section \ref{method} provides a description of the proposed IL framework and ISG. Section \ref{experiments} presents comprehensive experiments and an analysis of the results. Section \ref{conclusion} serves as our conclusion.

 \begin{figure*}[htp]
 \setlength{\abovecaptionskip}{0pt}
\centering
\includegraphics[width=1.0\linewidth]{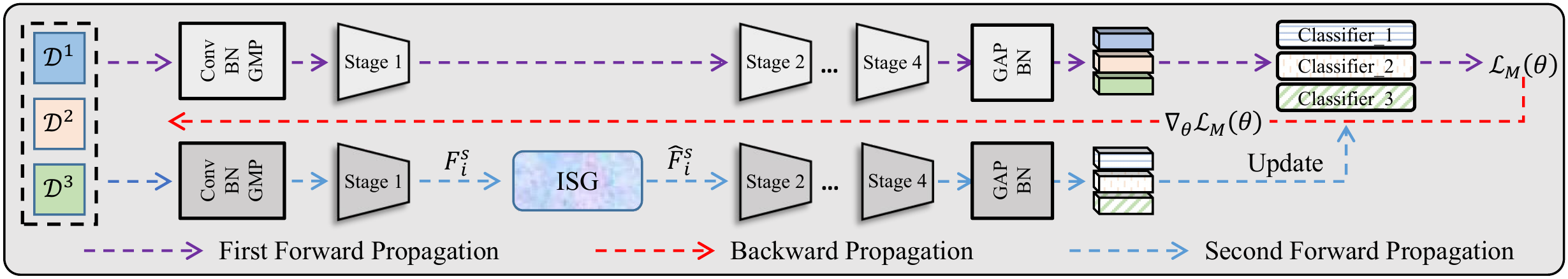}
\vspace{-1em}
\caption{
Illustration of the IL framework. We adopt the ResNet50 model as the backbone in this figure. Our framework incorporates two forward propagations and one backward propagation for each iteration. In the first forward propagation, we compute the loss $L_M(\theta)$ based on the features of the original styles and the class prototypes stored in the memory banks. In the backward propagation, parameters of the feature extractor are updated based on $ \nabla_{\theta}  \mathcal L_M(\theta)$. In the second forward propagation, we adopt the ISG module to generate stylized image features, which are used only to update memory banks. (Best viewed in color.)
}

\label{fig:net}
\vspace{-1em}
\end{figure*}

\section{Related Works} \label{related_work}
We divide our review of the related literature into two parts: 1) Domain Generalization (DG), and 2) Interleaved Learning (IL).

\subsection{Domain Generalization}
The goal of DG methods is to learn a model from one or several related source domains in a way that enables the model to generalize well to unseen target domains. Existing DG methods handle domain shift from various perspectives, including domain alignment \cite{shermin2020adversarial,muandet2013domain,motiian2017unified,li2020domain,li2018domain,li2018deep}, training strategy \cite{kim2022pin,balaji2018metareg,li2019feature}, data augmentation \cite{kang2022style,volpi2019addressing,shi2020towards,zhou2021domain,mancini2020towards}, and the causal mechanism \cite{mahajan2021domain,lv2022causality}. 

 Existing works in the field of ReID largely seek to improve DG performance from three perspectives: network architecture, training strategy, and data augmentation. For the first category of methods, 
Xu \textit{et al.} \cite{xu2022meta} designed a new network that learns both domain-specific and domain-invariant features. During testing, this network adaptively integrates the above features based on the correlation between the statistical information of the test sample and that of the source domains to produce more generalizable features. Jiao \textit{et al.} \cite{jiao2022dtin} introduced a new normalization module that employs dynamic convolution to remove styles while maintaining discriminative patterns during feature normalization. Choi \textit{et al.} \cite{choi2021meta} proposed a batch-instance normalization (BIN) module that combines batch normalization (BN) and instance normalization (IN). With the help of learnable balancing parameters, BIN can both reduce the style difference between domains and alleviate the loss of discriminative information. 

With regard to training strategies, some works adopt meta-learning \cite{li2018learning}. These works divide the source domains into multiple meta-train datasets and one meta-test dataset, which mimics the domain gap encountered during testing. Eliminating this domain gap during training can improve generalization ability. For instance, Zhang \textit{et al.} \cite{zhang2022acl} apply meta-learning to train a dynamic neural network capable of learning domain-specific features and embedding them into a shared feature space. As a result, the network is better able to extract generalizable features from an unknown target domain. Moreover, Zhao \textit{et al.} \cite{zhao2021learning} improved the traditional meta-learning by enriching the data distribution in the meta-test stage. These authors proposed a meta batch normalization (MetaBN) module that can inject domain information from the meta-train datasets into the meta-test features. This process generates a wider range of feature styles and improves the model's ability to generalize to new domains.

Finally, another popular method is data augmentation, which diversifies the styles of the source data and thereby improves the model's generalization ability. Most methods in this category directly change the feature statistics of one training image. For example, Nuriel \textit{et al.} \cite{nuriel2021permuted} proposed pAdaIN, which swaps feature statistics between samples in one batch. Zhou \textit{et al.} \cite{zhou2021domain} introduced MixStyle, an approach that combines the statistics of two samples in a linear manner. 
Li \textit{et al.} \cite{li2022uncertainty} designed DSU, which imposes disturbance on the original feature statistics. There are also methods that implement style transfer in the frequency domain. For example, Xu \textit{et al.} \cite{xu2021fourier} proposed the FACT method, which fuses or exchanges the low-frequency portions of the amplitude spectra of two images to achieve style transfer.

Our proposed IL approach belongs to the second category of methods in that it focuses on the training strategy. Unlike meta-learning, which imitates the domain gap by partitioning multiple source domains, IL introduces a domain gap between the feature extractor and classifier by synthesizing the features of interleaved styles. In the experimentation section, we demonstrate that IL is both more powerful and more efficient.

\subsection{Interleaved Learning} \label{text:example}
IL was first introduced in the fields of cognitive science and educational psychology \cite{halpern25learning,pashler2007organizing,carpenter2013effects}. In conventional learning, students are asked to complete exercises in a particular assignment in order to master a certain type of knowledge (for example, a dozen problems that can all be solved by using the Pythagorean theorem). This approach, which is referred to as “blocked learning”, results in students becoming aware of what kind of knowledge is required to solve each problem before they read the question. However, students that learn in this way may not perform well on a more comprehensive exam in which different types of problems are mixed together; in other words, the students ``overfit'' to the same problem type. In IL, each assignment includes different types of problems that are arranged in an interleaved order. Interleaved practice requires students to choose a strategy based on the problem itself rather than relying on a fixed strategy. Studies in cognitive science \cite{halpern25learning,pashler2007organizing,carpenter2013effects} have concluded that interleaving can effectively promote inductive learning.

Similar to the example of overfitting to the same problem-solving strategy described above, conventional ReID pipelines may result in overfitting to existing domain styles. To address this problem, we propose a novel IL framework for DG ReID. Our framework adopts the features of interleaved styles for classifier updating and loss computation, which prevents the feature extractor from overfitting to specific feature styles. Just as IL can help students to perform well when faced with various types of questions, it can also be used to efficiently improves the model’s generalization ability on unseen domains.

To the best of our knowledge, this is the first time that IL has been introduced to the field of ReID. Experimental results show that our framework significantly improves the DG ReID performance.

\section{Methodology} \label{method}
\label{Methodology}
	An overview of our style interleaved learning framework is presented in Fig.~\ref{fig:net}. For DG ReID, we are provided with $S$ source domains $\boldsymbol D_{source}=\{ \mathcal{D}^s\}^S_{s=1}$; here, $\mathcal{D}^s={({\bm x}^s_k,{ y}^s_k)}^{N^s}_{k=1}$. ${N^s}$ is the number of samples, while $S$ denotes the number of source domains. The label spaces of the source domains are disjoint. 
	The goal is to train a generalizable model using the source data. In the testing stage, the model is directly evaluated on the unseen target domain $\boldsymbol D_{target}$.

\subsection{Style Interleaved Learning Framework}
Our style interleaved learning framework (Fig. ~\ref{fig:net}) includes a CNN-based feature extractor $f_\theta$($\cdot$) and  maintains an individual memory-based classifier for each source domain.
    Unlike conventional learning, IL utilizes two forward propagations and one backward propagation for each iteration.

    In the first forward propagation, we do not artificially change the feature styles. Instead, feature vectors produced by $f_\theta$($\cdot$) are used to perform loss computations with class prototypes stored in memory banks (it should be noted that the memory banks remain unchanged in this step). In the backward propagation, moreover, the model is optimized in the same way as conventional learning strategies. In the second forward propagation, we introduce our proposed ISG, an effective method of generating stylized image features that are utilized to update the memory banks. For a source domain $\mathcal{D}^s$ with $K^s$ identities, its memory $\mathcal{M}^s$ has $K^s$ slots, where the $i$-th slot saves the prototype centroid $\boldsymbol{c}^s_{i}$ of the $i$-th identity. After this second forward propagation, no further backward propagation is required.

\paragraph{The First Forward Propagation}

During each training iteration, for an image $\bm{x}^s_i$ from $\mathcal{D}^s$,  we forward it through the feature extractor
and obtain the L2-normalized  feature $\boldsymbol{f}^s_i$, \textit{i.e.}, $\boldsymbol{f}^s_i$ = $f_\theta$($\bm{x}^s_i$). We calculate the memory-based identification loss as follows:
	\begin{equation}\label{eq:loss_contrastive}
    \mathcal{L}_s = -\sum_{i=1}^{N^s} \log\frac{exp(\langle \boldsymbol{f}^s_i, \boldsymbol{c^s_{+}} \rangle /\tau)}{\sum_{k=1}^{K^s}{exp(\langle \boldsymbol{f}^s_i, \boldsymbol{c}^s_k \rangle /\tau)}},
\end{equation}
where $\boldsymbol{c^s_{+}}$ stands for the positive class prototype corresponding to $\boldsymbol{f}^s_i$, $\tau$ is the temperature factor, and $\langle \cdot , \cdot  \rangle$ denotes the computation of cosine similarity. 
The loss value is low when $\boldsymbol{f}^s_i$ is similar to $\boldsymbol{c^s_{+}}$ and dissimilar to all other class prototypes. It is worth noting that $\boldsymbol{f}^s_i$ is not used to update the memory bank.

\paragraph{The Backward Propagation}

The total loss is a combination of identification losses on all source domains, which is used to optimize the model via gradient descent:
\begin{equation}\label{eq:whole loss}
\mathcal L_M(\theta) = \frac{1}{S} \sum_{s=1}^{S} \mathcal{L}_s, ~~~~
\end{equation}
	\begin{equation}\label{eq:update feature extractor}
\theta'\leftarrow \theta - \alpha \nabla_{\theta}  \mathcal L_M(\theta),
\end{equation}
where $\theta$ denotes the parameters of $f_\theta$($\cdot$), while $\alpha$ is the learning rate.

\paragraph{The Second Forward Propagation}

The core concept of IL involves adopting the features of different styles for memory updating and loss computation. The styles generated in the second forward pass should be interleaved with the sample styles from the first forward pass while still ensuring that the semantic content of the image is retained. To achieve this goal, we introduce an ISG module to transform the feature styles.
	 
	 In more detail, we denote the feature maps of $\bm{x}_i^s$ output by a certain layer of  $f_{{\theta}^{'}}$($\cdot$) as $\bm{F}^s_i \in \mathbb{R}^{C \times H \times W}$, where $C$, $H$, and $W$ denote the number of channels, the height, and the width, respectively. We transform the styles of $\bm{F}^s_i$ in the following way:
	\begin{equation} \label{eq:AS}
\bm{\hat{F}}^s_i = \operatorname{ISG}(\bm{F}^s_i),
\end{equation}
where $\operatorname{ISG}(\cdot)$ is a feature stylization approach, the details of which will be introduced in Section \ref{FS}.

We next forward $\bm{\hat{F}}^s_i$ through the remaining layers of $f_{{\theta}^{'}}$($\cdot$) and obtain the L2-normalized feature vector $\bm{\hat{f}}^s_i$. In each iteration, we adopt $\bm{\hat{f}}^s_i$ to update the corresponding class prototype $\boldsymbol{c}^s_{+}$ in the memory banks:	
		\begin{equation}
\label{eq:update}
    \boldsymbol{c}^s_{+} \gets \eta \boldsymbol{c}^s_{+} + (1-\eta)\bm{\hat{f}}^s_i, ~~~\bm{\hat{f}}^s_i \in {\cal I}_{+},
\end{equation}
	where $\eta \in [0,1]$ is a momentum coefficient, while ${\cal I}_{+}$ denotes the set of samples belonging to the identity of $\bm{x}^s_i$ in the batch. Our style interleaved learning framework repeats the above three steps until training is complete. It should be noted that the second forward propagation is highly efficient and introduces only a small additional computational cost; more details will be provided in Section \ref{complex}.

% \subsection{Reliable Sample Selection by Collaborative Clustering}
\subsection{Interleaved Style Generator} \label{FS}

% \vspace{-5pt}
\paragraph{Preliminaries}
 Recent studies on style transfer \cite{huang2017arbitrary,zhou2021domain,li2022uncertainty} suggest that the style information for each image can be revealed by the feature statistics of one CNN bottom layer. Specifically, for a feature map $\bm{F} \in \mathbb{R}^{C \times H\times W}$, its style can be represented via $\mu(\bm{F}), \sigma(\bm{F}) \in \mathbb{R}^C$, which respectively store the means and standard deviations computed within each channel of $\bm{F}$:
\begin{equation} \label{eq:instance_mean}
\mu_c(\bm{F}) = \frac{1}{HW} \sum_{h=1}^H \sum_{w=1}^W \bm{F}_{chw},
\end{equation}

\begin{equation} \label{eq:instance_std}
\sigma_c(\bm{F}) = \sqrt{ \frac{1}{HW} \sum_{h=1}^H \sum_{w=1}^W ( \bm{F}_{chw} - \mu_c(\bm{F}) )^2 }.
\end{equation}

\begin{figure*}
\setlength{\abovecaptionskip}{0pt}
\centering
\includegraphics[width=1.0\linewidth]{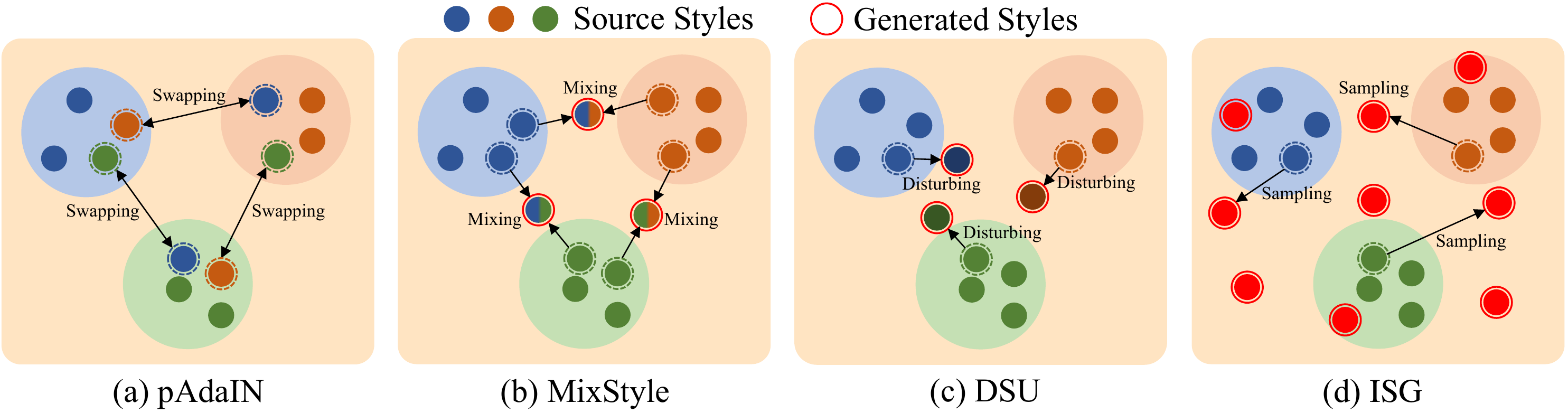}

\caption{
Comparisons between ISG and pAdaIN \cite{nuriel2021permuted}, MixStyle \cite{zhou2021domain}, and DSU \cite{li2022uncertainty}. We assume there are three source domains in this figure. (a) pAdaIN swaps feature styles between two samples. (b) MixStyle linearly mixes the styles of two samples. (c) DSU imposes disturbance on the original style of each sample. (d) ISG uniformly samples a style within the estimated style space. Best viewed in color.
}

\label{fig:visual}
\vspace{-1em}
\end{figure*}

It is therefore reasonable to change the feature style of an image by modifying its feature statistics. The obtained new feature map  can be represented as follows:
\begin{equation} \label{eq:style_transfer}
\hat{\bm{F}} = \bm{\gamma} \odot \frac{\bm{F} - \mu(\bm{F})}{\sigma(\bm{F})} + \bm{\beta},
\end{equation}
where $\bm{\gamma}$ and $\bm{\beta}$ denote the channel-wise affine transformation parameters, while $\odot$ is the Hadamard product. Various approaches have been proposed to obtain reasonable values of $\bm{\gamma}$ and $\bm{\beta}$. As illustrated in Fig.  \ref{fig:visual}, pAdaIN \cite{nuriel2021permuted} swaps the feature statistics of two images in one batch, with the newly introduced values defined as the $\bm{\gamma}$ and $\bm{\beta}$ for each image. MixStyle \cite{zhou2021domain} mixes the feature statistics of two images in a linear manner to obtain $\bm{\gamma}$ and $\bm{\beta}$. DSU \cite{li2022uncertainty} imposes disturbances on the original feature statistics of each image to obtain the affine transformation parameters.

\paragraph{Proposed Approach}

The IL framework favors  interleaved styles.
However, the feature styles synthesized by most existing approaches \cite{zhou2021domain,li2022uncertainty} remain closely related to those of the training images, which deviates from the spirit of IL. In the following, we propose an Interleaved Style Generator, which generates styles that are both meaningful and independent from the original styles in each batch.

Similar to \cite{li2022uncertainty}, we model the distributions of both style vectors in Eq.~\ref{eq:instance_mean} and Eq.~\ref{eq:instance_std} using Gaussian distributions. 
 For the sake of simplicity, we denote the mean and standard deviation vectors for the $b$-th instance in a batch as $\bm{\mu}_{b} \in \mathbb{R}^C$ and $\bm{\sigma}_{b} \in \mathbb{R}^C$ respectively. We first calculate the distributions of both style vectors in a mini-batch as follows: 
\begin{equation}
\label{equ:channelwise_statistics}
    \hat{\bm{\mu}}_{\mu} = \frac{1}{B}\sum_{b=1}^{B} \bm{\mu}_b, ~~
    \hat{\bm{\mu}}_{\sigma} = \frac{1}{B}\sum_{b=1}^{B} \bm{\sigma}_b,
\end{equation}

\begin{equation}
    \label{equ:channelwise_statistics}
    \begin{split}
        \hat{\bm{\sigma}}^{2}_{\mu} =\frac{1}{B}\sum_{b=1}^{B} (\bm{\mu}_b - \hat{\bm{\mu}}_{\mu})\odot (\bm{\mu}_b - \hat{\bm{\mu}}_{\mu}) , \\
        \hat{\bm{\sigma}}^{2}_{\sigma} = \frac{1}{B}\sum_{b=1}^{B} (\bm{\sigma}_b - \hat{\bm{\mu}}_{\sigma})\odot (\bm{\sigma}_b - \hat{\bm{\mu}}_{\sigma}),
    \end{split}
\end{equation}
where $\hat{\bm{\mu}}_{\mu}$ and $\hat{\bm{\sigma}}_{\mu}$ characterize the distribution of the style vector in Eq.~\ref{eq:instance_mean}, while $\hat{\bm{\mu}}_{\sigma}$ and $\hat{\bm{\sigma}}_{\sigma}$ describe the distribution of the style vector in Eq.~\ref{eq:instance_std} and B is the batch size.

Subsequently, we represent the range of meaningful feature styles as  $[\hat{\bm{\mu}}_{\mu}-\rho*\hat{\bm{\sigma}}_{\mu},\hat{\bm{\mu}}_{\mu}+\rho*\hat{\bm{\sigma}}_{\mu}]$ and $[\hat{\bm{\mu}}_{\sigma}-\rho*\hat{\bm{\sigma}}_{\sigma},\hat{\bm{\mu}}_{\sigma}+\rho*\hat{\bm{\sigma}}_{\sigma}]$, where $\rho$ is a hyper-parameter. We then obtain new feature styles via uniform sampling within the above intervals, as follows:
\begin{align}
  \bm{\beta}_{\text{ISG}} \sim U_{\mu}(\hat{\bm{\mu}}_{\mu}-\rho*\hat{\bm{\sigma}}_{\mu}, \hat{\bm{\mu}}_{\mu}+\rho*\hat{\bm{\sigma}}_{\mu}),
  \label{equ:sample_new_style1}\\
  \bm{\gamma}_{\text{ISG}} \sim U_{\sigma}(\hat{\bm{\mu}}_{\sigma}-\rho*\hat{\bm{\sigma}}_{\sigma},\hat{\bm{\mu}}_{\sigma}+\rho*\hat{\bm{\sigma}}_{\sigma}).
  \label{equ:sample_new_style2}
  \end{align}

Finally, we modify the style of $\bm{F}$ by replacing the $\bm{\gamma}$ and $\bm{\beta}$ in Eq.~\ref{eq:style_transfer} with $\bm{\gamma}_{\text{ISG}}$ and $\bm{\beta}_{\text{ISG}}$ to achieve meaningful style transfer, as follows:
\begin{equation} \label{eq:mixstyle}
{ISG}(\bm{F}) = \bm{\gamma}_{\text{ISG}} \odot \frac{\bm{F} - \mu(\bm{F})}{\sigma(\bm{F})} + \bm{\beta}_{\text{ISG}}.
\end{equation}

The use of uniform sampling has two key benefits. First, the obtained $\bm{\beta}_{\text{ISG}}$ and $\bm{\gamma}_{\text{ISG}}$  are independent from the original styles $\mu(\bm{F})$ and $\sigma(\bm{F})$.
Therefore, ISG is highly suitable for use with IL, which requires features of interleaved styles for the updating of the feature extractor and classifiers. Second, as revealed in Fig.  \ref{fig:visual}, the obtained styles are more diverse than those produced by existing works such as \cite{nuriel2021permuted,zhou2021domain,li2022uncertainty}.  

We insert the ISG module after one bottom CNN layer (\textit{e.g.}, at the first stage of the ResNet50 model).
Since ISG is parameter-free and is used only in the second forward propagation, the computational cost it introduces is very small. During inference, we remove ISG from the feature extractor. A pseudo-code for ISG is presented in Algorithm \ref{alg:mixstyle_pytorch}.

\begin{algorithm}[h]
\caption{Pseudo-code for Interleaved Style Generator.}
\label{alg:mixstyle_pytorch}
\definecolor{codeblue}{rgb}{0.25,0.5,0.5}
\lstset{
  backgroundcolor=\color{white},
  basicstyle=\fontsize{7.2pt}{7.2pt}\ttfamily\selectfont,
  columns=fullflexible,
  breaklines=true,
  captionpos=b,
  commentstyle=\fontsize{7.2pt}{7.2pt}\color{codeblue},
  keywordstyle=\fontsize{7.2pt}{7.2pt},
}
\begin{lstlisting}[language=python]
# x: input features of shape (B, C, H, W)
# p: probabillity to apply ISG (default: 1.0)
# rho: hyper-parameter for the range of uniform distribution (default: 3.0)
# eps: a small value added before square root for numerical stability (default: 1e-6)

if not in training mode:
    return x

if random probability > p:
    return x

B,C= x.size(0),x.size(1)

# compute instance mean & standard deviation
mu = x.mean(dim=[2, 3], keepdim=True) 
var = x.var(dim=[2, 3], keepdim=True) 
sig = (var + eps).sqrt() 

# normalize input
x_normed = (x - mu) / sig 

# compute the parameter of two intervals
mu_mu = mu.mean(0).squeeze()
sig_mu = sig.mean(0).squeeze() 

mu_std = mu.std(0).squeeze()
sig_std = sig.std(0).squeeze()

# perform uniform sampling 
Uniform_mu = Uniform(low=mu_mu - rho*mu_std, 
                    high=mu_mu + rho*mu_std)
Unifrom_sig = Uniform(low=sig_mu - rho*sig_std, 
                    high=sig_mu + rho*sig_std)

mu_isg = Uniform_mu.sample([B, ]).reshape([B,C,1,1])
sig_isg = Uniform_sig.sample([B, ]).reshape([B,C,1,1])

# denormalize input using the sampled statistics
return x_normed * sig_isg + mu_isg 
\end{lstlisting}
\end{algorithm}

\section{Experiments} \label{experiments}

\subsection{Datasets and Settings}

\textbf{Datasets.}
We conduct extensive experiments on several public ReID datasets, namely Market1501 \cite{zheng2015scalable}, DukeMTMC-ReID \cite{zheng2017unlabeled}, CUHK03 \cite{li2014deepreid}, MSMT17 \cite{wei2018person} and CUHK-SYSU \cite{xiao2016end}.
It is worth noting that DukeMTMC-reID has been widely adopted in existing DG ReID works \cite{jiao2022dtin,choi2021meta,zhao2021learning,dai2021generalizable,zhou2021learning,liao2020interpretable,jin2020style}. In particular, to the best of our knowledge, there is no alternative setting for the single-source DG ReID task to Protocol-3 introduced below. Therefore, this database is still adopted in this work. For simplicity, we abbreviate the names of these datasets as M, D, C3, MS and CS, respectively. Adopting the same approach as \cite{dai2021generalizable,choi2021meta,yu2021multiple}, all images in each source dataset are used for training, regardless of the train or test splits provided in the individual protocols.
We adopt mean average precision (mAP) and Rank-1 accuracy as the evaluation metrics.
\begin{table*}[htp]
\centering
    \caption{
    Comparisons with state-of-the-art methods on multi-source DG ReID  benchmarks under Protocol-1. $\dag$ indicates evaluation results based on the code released by the authors.}
    \label{tab:sota}
    \vspace{-1em}
    \resizebox{0.95\textwidth}{!}{
        \begin{tabular}{c||c|cc|cc|cc|cc|cc}
            \toprule[1pt]
            \multirow{2}{*}{\textbf{Method}}&\multirow{2}{*}{\textbf{Backbone}} &\multicolumn{2}{c|}{D+C3+MS$\to$ M}&\multicolumn{2}{c|}{M+C3+MS$\to$D}&\multicolumn{2}{c|}{M+D+C3$\to$MS}&\multicolumn{2}{c|}{M+D+MS$\to$C3}& \multicolumn{2}{c}{\textbf{Average}}\\ \cline{3-12}
            &&~mAP~&Rank-1 &~mAP~&Rank-1&~mAP~&Rank-1&~mAP~&Rank-1&~mAP~&Rank-1
            \\ \hline\hline
            QAConv \cite{liao2020interpretable}&ResNet50&35.6&65.7&47.1&66.1&7.5&24.3&21.0&23.5&27.8&44.9\\
            CBN \cite{zhuang2020rethinking}&ResNet50&47.3&74.7&50.1&70.0&15.4&37.0&25.7&25.2&34.6&51.7\\
            SNR \cite{jin2020style}&SNR50& 48.5&75.2&48.3&66.7&13.8&35.1&29.0&29.1&34.9 & 51.5\\
            OSNet \cite{zhou2021learning}&OSNet& 44.2&72.5&47.0&65.2&12.6&33.2&23.3&23.9& 31.8 & 48.7\\
            OSNet-IBN \cite{zhou2021learning}&OSNet-IBN& 44.9&73.0&45.7&64.6&16.2&39.8&25.4&25.7& 33.0 & 50.8\\
            OSNet-AIN \cite{zhou2021learning}&OSNet-AIN& 45.8&73.3&47.2&65.6&16.2&40.2&27.1&27.4& 34.1 & 51.6\\
             MECL \cite{yu2021multiple}&ResNet50&56.5&80.0&53.4&70.0&13.3&32.7&31.5&32.1& 38.7 & 53.7 \\
             RaMoE \cite{dai2021generalizable}&ResNet50  & 56.5   & 82.0 & 56.9  & 73.6  & 13.5   & 34.1  & 35.5 & 36.6  &40.6  &56.6  \\
             MixNorm \cite{qi2022novel} &ResNet50& 51.4 & 78.9 &49.9 &70.8 &19.4 &47.2&29.0 &29.6&37.4 &56.6\\ 
             M$^3$L$^\dag$ \cite{zhao2021learning} &ResNet50& 59.6 & 81.5 &54.5 &71.8 &16.0 &36.9&35.2 &36.4&41.3 &56.7 \\
             MetaBIN$^\dag$ \cite{choi2021meta} &ResNet50-BIN& 61.2 & 83.2 &54.9 &71.3 &17.0 &40.8&37.5 &38.1&42.7 &58.4\\
             META$^\dag$ \cite{xu2022meta} &META &65.7 &85.3  &59.9&76.9 &22.5&49.3 & 45.9&46.0 &48.5&64.4  \\ 
             ACL$^\dag$ \cite{zhang2022acl}  &ACL &70.1&88.4&60.6&75.9&23.5&51.9 &47.7 &48.0 &50.5 &66.1    \\
             \hline
            Baseline &ResNet50     &59.3      & 81.2        & 54.3 & 70.9 &  14.7 &  35.2       &36.1  &37.4 &41.1 &56.2 \\
            IL  &ResNet50     &65.8    & 86.2        & 57.1 & 75.4 & 20.2    &  45.7       &38.3  &40.9 &45.4& 62.1 \\
            IL  &META  &68.9 &86.4 &60.8 &77.1 &24.1&52.0& \textbf{48.0}& \textbf{48.2} &50.5&65.9   \\
            Baseline &ACL  &68.5&87.1  &58.4&75.9  &20.7&47.1  &44.0&45.2 &47.9&63.8   \\
            IL  &ACL & \textbf{72.5} & \textbf{89.7}  & \textbf{61.9} & \textbf{78.2}  & \textbf{25.7} & \textbf{54.0}  &46.7 &46.9 & \textbf{51.7}& \textbf{67.2} \\
            
            \toprule[1pt]
        \end{tabular}
    }
\vspace{-1em}
\end{table*}

\begin{table*}[htp]
\centering
\caption{
    Comparisons with state-of-the-art methods on multi-source DG ReID  benchmarks under Protocol-2.}
\label{tab:sota2}
\vspace{-1em}
\resizebox{0.95\linewidth}{!}{
\begin{tabular}{c||c|cc|cc|cc|cc}
\toprule[1pt]
\multirow{2}{*}{\textbf{Method}} &
%   \multirow{2}{*}{\textbf{\begin{tabular}[c]{@{}c@{}}Backbone \\ Params\end{tabular}}} &
\multirow{2}{*}{\textbf{Backbone}} &
  \multicolumn{2}{c|}{M+MS+CS$\to$C3} &
  \multicolumn{2}{c|}{M+CS+C3$\to$MS} &
  \multicolumn{2}{c|}{MS+CS+C3$\to$M} &
  \multicolumn{2}{c}{\textbf{Average}} \\ \cline{3-10} 
        &              & mAP  & Rank-1 & mAP  & Rank-1 & mAP  & Rank-1 & mAP  & Rank-1 \\ \hline \hline
SNR \cite{jin2020style}     & SNR50        & 17.5 & 17.1   & 7.7  & 22.0   & 52.4 & 77.8   & 25.9 & 39.0   \\
QAConv \cite{liao2020interpretable}  & QAConv50     & 32.9 & 33.3   & 17.6 & 46.6   & 66.5 & 85.0   & 39.0 & 55.0   \\
M$^3$L \cite{zhao2021learning}    & ResNet50     & 32.3 & 33.8   & 16.2 & 36.9   & 61.2 & 81.2   & 36.6 & 50.6   \\
M$^3$L \cite{zhao2021learning} & ResNet50-IBN & 35.7 & 36.5   & 17.4 & 38.6   & 62.4 & 82.7   & 38.5 & 52.6   \\
MetaBIN \cite{choi2021meta} & ResNet50-BIN & 43.0 & 43.1   & 18.8 & 41.2   & 67.2 & 84.5   & 43.0 & 56.3   \\
ACL \cite{zhang2022acl}     & ACL      & \textbf{49.4} & \textbf{50.1}   & 21.7 & 47.3   & 76.8 & 90.6   & 49.3 & 62.7   \\
META \cite{xu2022meta}    & META        & 47.1 & 46.2   & 24.4 & 52.1   & 76.5 & 90.5   & 49.3 & 62.9   \\ \hline
Baseline      & ResNet50     &35.6  &36.1    &17.6  &38.0    &66.4  &84.6    &39.9  &52.9    \\
IL      & ResNet50     & 41.0 & 41.8   & 23.8 & 51.2   & 72.0 & 88.5   & 45.6 & 60.5   \\
Baseline      & ACL     &41.9  &42.3    &21.4  &44.1    &72.4  &87.9    &45.2  &58.1    \\
IL & ACL            & 47.6 & 48.3   & 26.8 & \textbf{54.8}   & 78.2 & 90.7   & 50.9 & 64.6   \\ 
IL & META             & 48.9 & 48.8   & \textbf{26.9} & \textbf{54.8}   & \textbf{78.9} & \textbf{91.2}   & \textbf{51.6} & \textbf{64.9}   \\ \toprule[1pt]
\end{tabular}}
\vspace{-1em}
\end{table*}

\textbf{Settings.} To facilitate comprehensive comparisons with existing works \cite{dai2021generalizable,choi2021meta,jin2020style}, we adopt three popular evaluation protocols.

\textbf{Protocol-1.} 
This is the leave-one-out setting for M, D, C3, and MS. This setting selects one dataset from the four for testing and uses the remaining datasets for training. 

\textbf{Protocol-2.} 
This is the leave-one-out setting for M, C3, and MS. This setting selects one dataset from the three for testing and uses the remaining datasets plus CS for training.

\textbf{Protocol-3.} This protocol includes the M and D datasets, which take turns being used as the source domain and target domain.

\subsection{Implementation Details}
 We use the ResNet50 model \cite{he2016deep} pretrained on ImageNet \cite{deng2009imagenet} as the feature extractor. Following \cite{luo2019bag,dai2021generalizable,zhao2021learning}, we set the stride of the last residual layer as 1. We sample 64 images from each source domain, including 16 identities and 4 images per identity; as a result, our batch size is 64 $\times$ $S$. For data augmentation, we perform random cropping and random flipping. For the memory, $\eta$ and $\tau$ are set to 0.2 and 0.05, in line with \cite{zhao2021learning}. For the Batch-Style sampler, $\rho$ is set to 3. We optimize the model using the Adam optimizer and train the model for 70 epochs. The learning rate is initialized as 3.5 $\times$ 10$^{-4}$ and then divided by 10 at the 30-th and 50-th epochs. We use the warmup strategy \cite{luo2019bag} in the first 10 epochs. Finally, we adopt the same ResNet50 model optimized according to the conventional training strategy (Fig. ~\ref{fig:example} (a)) as the baseline. The augmentation setting (Aug) mentioned in this section  incorporates one forward pass and one backward pass. It employs one feature stylization method in the forward pass and adopts the output feature for loss computation and classifier updating simultaneously. In comparison, our IL framework consists of two forward passes and one backward pass. In the first forward pass, we do not use any feature stylization methods and do not update the memory-based classifiers. In the second forward pass, we employ the feature stylization method and use the stylised features to update the classifiers.

%-------------------------------------------------------------------------

\subsection{Comparisons with State-of-the-Art Methods} \label{Comparison sota}

\textbf{Protocol-1}. To facilitate fair comparison, we adopt the same training data as in \cite{dai2021generalizable} to \cite{zhao2021learning,choi2021meta} and obtain better results than those reported in the original paper. The comparisons in Table \ref{tab:sota} show that our method consistently outperforms state-of-the-art methods by notable margins. In particular, our method outperforms comparison methods based on meta-learning (\textit{e.g.},  RaMoE \cite{dai2021generalizable}, M$^3$L \cite{zhao2021learning}, and MetaBIN \cite{choi2021meta}).

The interleaved and meta-learning strategies solve the DG ReID problem from different perspectives. Specifically, in interleaved learning, the styles of the features used for classifier updating are different from those used for loss computation; this prevents the feature extractor from overfitting to the feature styles contained in the source domain data. In comparison, meta-learning divides the source domains into meta-train and meta-test domains to simulate the domain shift that will be encountered during the testing stage. However, the classifier for each domain is still updated according to the same features as those used for loss computation, which has a risk of overfitting to source domains.

\begin{table}[tp]
\caption{Comparisons with state-of-the-art methods on single-source DG ReID benchmarks under Protocol-3.}
\label{tab:single}
\vspace{-1em}
\centering
\resizebox{0.475\textwidth}{!}{%
\begin{tabular}{c||c|cc|cc}
\toprule[1pt]
\multirow{2}{*}{\textbf{Method}} & \multirow{2}{*}{\textbf{Backbone}}& \multicolumn{2}{c|}{M$\to$D} & \multicolumn{2}{c}{D$\to$M}   \\ \cline{3-6} 
& & mAP  & Rank-1& mAP  & Rank-1  \\ \hline \hline
IBN-Net \cite{pan2018two}&IBN-Net& 24.3 & 43.7  & 23.5& 50.7  \\
OSNet \cite{zhou2021learning}&OSNet& 25.9& 44.7   & 24.0 & 52.2  \\
OSNet-IBN \cite{zhou2021learning}&OSNet-IBN& 27.6& 47.9    & 27.4 & 57.8 \\
CrossGrad \cite{shankar2018generalizing}&ResNet50& 27.1& 48.5   & 26.3 & 56.7  \\
QAConv \cite{liao2020interpretable}&QAConv50& 28.7& 48.8 & 27.2& 58.6 \\
L2A-OT \cite{zhou2020learning}&ResNet50& 29.2& 50.1   & 30.2& 63.8   \\
OSNet-AIN \cite{zhou2021learning}&OSNet-AIN& 30.5& 52.4  & 30.6 & 61.0  \\
SNR \cite{jin2020style} &SNR50&33.6& 55.1 & 33.9& 66.7  \\
MetaBIN \cite{choi2021meta}&ResNet50-BIN& 33.1& 55.2    & 35.9 & 69.2 \\ 
DTIN-Net \cite{jiao2022dtin}&ResNet50-DTIN& \textbf{36.1}& 57.0    & 37.4 & \textbf{69.8} \\ \hline
Baseline &ResNet50&  31.4& 50.1  & 31.2& 59.6   \\
% IL  &ResNet50&  34.7 & 55.5  & 39.5& 68.9 \\
IL  &ResNet50&  35.0 & \textbf{57.4}  & \textbf{39.9}& \textbf{69.8} \\
\toprule[1pt]
\end{tabular}}
\vspace{-1em}
\end{table}

\textbf{Protocol-2.}
As shown in Table \ref{tab:sota2}, IL achieves superior performance with the ResNet50 backbone. Moreover, IL can be readily applied to stronger backbones such as ACL \cite{zhang2022acl} or META \cite{xu2022meta}. For example, we simply replace the ResNet50 with an ACL backbone~\cite{zhang2022acl}. It should be noted that the training strategies adopted in~\cite{zhang2022acl} (\textit{e.g.}, meta-learning and cluster loss) are not employed in our experiments. Instead, we  adopt the cross-entropy loss function $\mathcal{L}_s$ (Eq.~\ref{eq:loss_contrastive}). It can be observed that IL outperforms both ACL and our baseline. Moreover, META consists of a global branch and $S$ expert branches. We apply IL to the global branch only, leaving the expert branches unchanged to facilitate the learning of domain-specific knowledge. It is clear that the combination of IL and META yields superior performance. This experiment demonstrates the flexibility of IL as a plug-and-play strategy that can be used with other DG methods or backbones. 

\textbf{Protocol-3.}
As shown in Table \ref{tab:single}, our method still outperforms most state-of-the-art methods under Protocol-3. Some recent works have proposed stronger model architectures for DG ReID. For example, SNR \cite{jin2020style}, MetaBIN \cite{choi2021meta} and DTIN-Net \cite{jiao2022dtin} attempt to eliminate style discrepancies between instances by inserting IN layers into the backbone model. Our method does not change the model structure during inference and achieves better performance. Moreover, our style interleaved learning framework significantly improves the performance relative to the baseline. The above comparisons demonstrate the effectiveness of the IL strategy.

% %-------------------------------------------------------------------------

\subsection{Ablation Study}

To verify the effectiveness of each component in our IL framework, we conduct an ablation study under Protocol-1. 

\textbf{Interleaved learning framework.} 
As is evident from Table \ref{tab:ablation study}, IL achieves significantly better performance than the baseline. This is because interleaved feature styles introduce a domain shift between the feature extractor and classifiers. Eliminating this domain shift improves the generalization ability of the feature extractor. 

\begin{figure}[tp]

\setlength{\abovecaptionskip}{0pt}
\centering
\includegraphics[width=\linewidth]{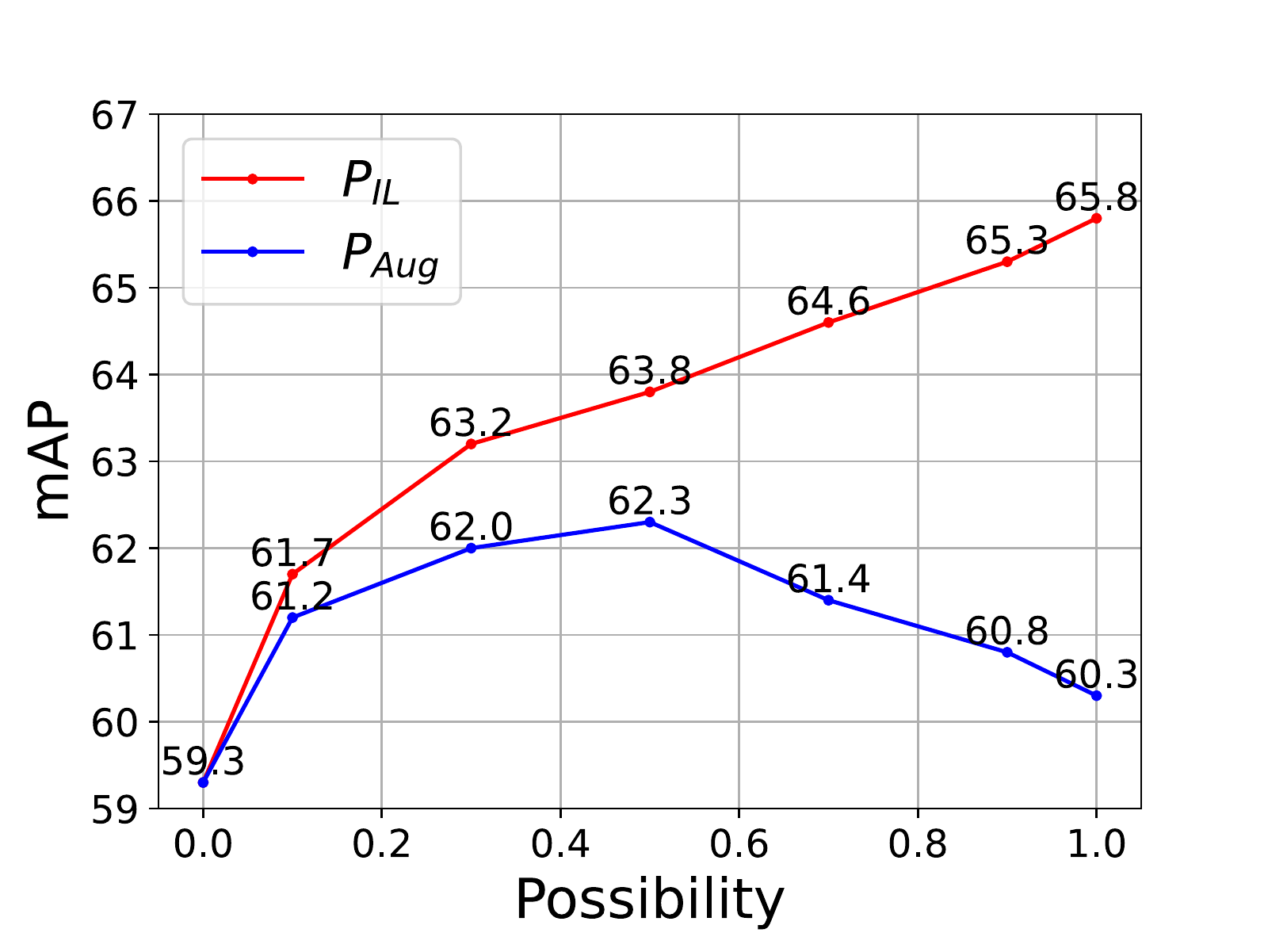}
\caption{Performance comparison between IL and data augmentation. $P_{IL}$ and $P_{Aug}$ denote the probability of ISG being activated under the IL framework and the common data augmentation setting, respectively. Experiments are conducted under the D+C3+MS$\to$M setting.}
\label{fig:compare_PIL_Paug}
\end{figure}

Moreover, as shown in Table \ref{tab:ablation study} and Fig.  \ref{fig:compare_PIL_Paug}, IL outperforms the common data augmentation strategy, which involves utilizing ISG to diversify feature styles for both the feature extractor and the classifiers. When ISG is employed for data augmentation, the best activation probability is 0.5. In comparison, when it is used in the second forward pass of IL, the model generalization ability consistently improves as the activation probability increases. The above experimental results demonstrate the superiority of the IL framework.

\begin{table}[tp]

\centering
\caption{
Ablation study on each key component. A tick next to ``ISG” only indicates that ISG is utilized for data augmentation in the conventional learning strategy. A tick next to ``IL” means that the IL scheme is adopted.
}
\vspace{-1em}
\label{tab:ablation study}
\begin{tabular}{cc||cc|cc}
% \hline
\toprule[1pt]
 \multirow{2}{*}{ISG} & \multirow{2}{*}{IL} & \multicolumn{2}{c|}{D+C3+MS$\to$M} & \multicolumn{2}{c}{M+D+C3$\to$MS} \\ \cline{3-6} 
 & & mAP & Rank-1 & mAP & Rank-1    \\ 
 \hline \hline
& & 59.3 & 81.2 &  14.7 &  35.2 \\ 
$\checkmark$  &  &62.3  &83.3  &15.3  &36.9  \\
$\checkmark$  &$\checkmark$& 65.8 &86.2  &20.2  &45.7 \\
 \toprule[1pt]
\end{tabular}
\vspace{-1em}
\end{table}

\begin{table}[tp]
\caption{Comparisons between ISG and existing feature stylization methods. Experiments are conducted under the D+C3+MS$\to$M setting.}
\label{tab:AUGvsIL}
\vspace{-1em}
\resizebox{0.475\textwidth}{!}{%
\begin{tabular}{c||cc|cc|cc}
\toprule[1pt]
\multirow{2}{*}{Method} & \multicolumn{2}{c|}{MixStyle} & \multicolumn{2}{c|}{DSU} & \multicolumn{2}{c}{ISG} \\ \cline{2-7} 
    & mAP  & Rank-1 & mAP  & Rank-1 & mAP  & Rank-1 \\ \hline \hline
Baseline &59.3& 81.2& 59.3 & 81.2   & 59.3 & 81.2   \\ \hline
Augment & 60.9 & 82.3   & 61.3 & 82.5   & 62.3 & 83.3   \\
IL  & 62.7 & 83.2   & 62.1 & 83.1   & 65.8 & 86.2   \\ \toprule[1pt]
\end{tabular}}
\vspace{-1em}
\end{table}

\textbf{Comparison with existing stylization methods.} 
We conduct comparisons with two representative stylization methods, named MixStyle \cite{zhou2021domain} and DSU \cite{li2022uncertainty}, and present the results in Table \ref{tab:AUGvsIL}. We apply each of these methods to both the data augmentation and IL strategies. 

From these results, we can draw two conclusions: 1) Feature stylization methods perform better when applied to the interleaved learning framework. This suggests that updating the feature extractors and classifiers according to the features of interleaved styles is indeed beneficial. 2) ISG performs significantly better than MixStyle and DSU in the IL setting. Combined with the analysis in Fig.  \ref{fig:visual}, we can safely conclude that synthesizing feature styles that are independent of the original styles is an essential element of successful IL.

\begin{figure}
\setlength{\abovecaptionskip}{0pt}
\centering
\includegraphics[width=\linewidth]{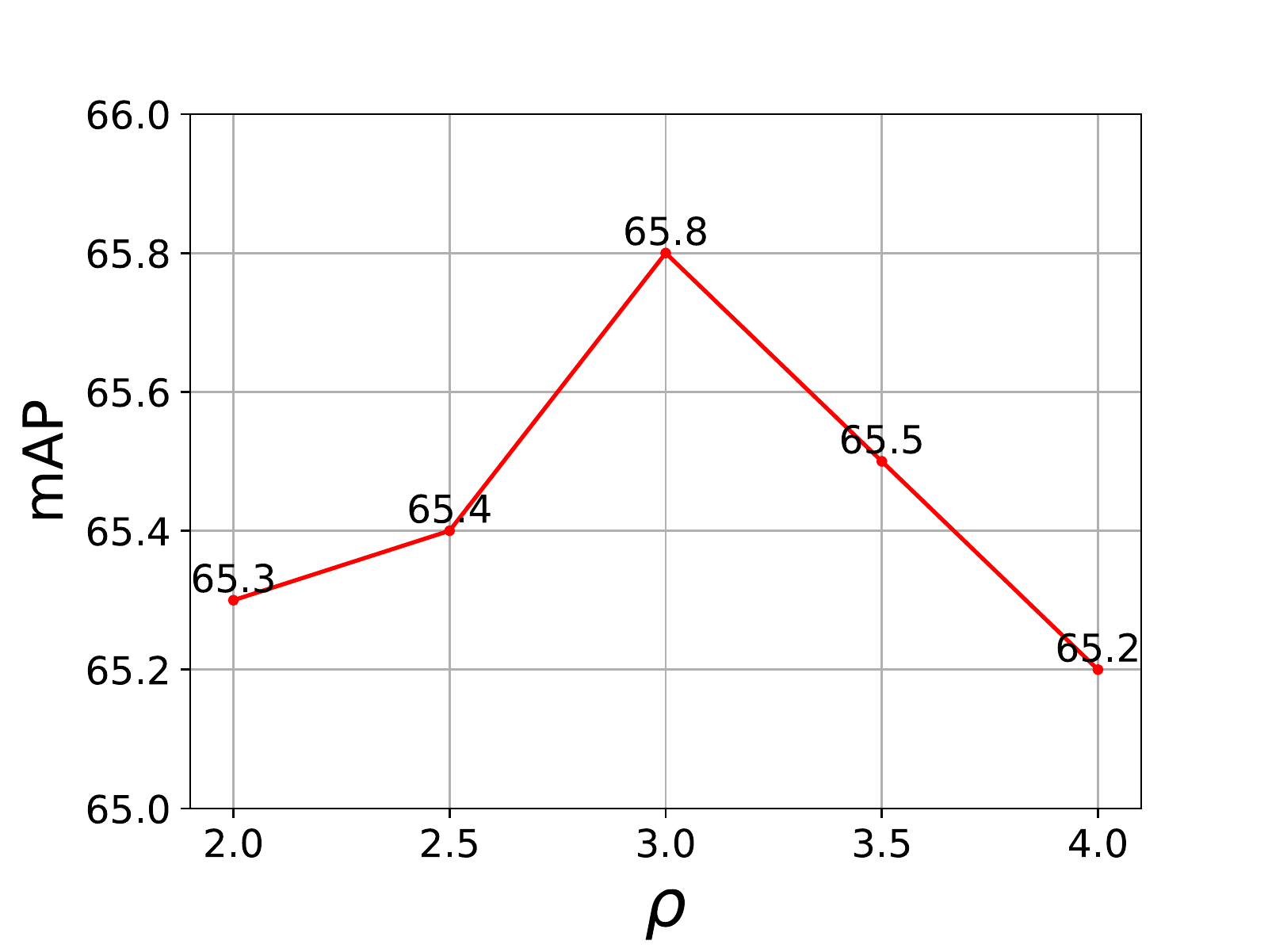}
\caption{Ablation study on the value of $\rho$. Experiments are conducted under the D+C3+MS$\to$M setting.}
\label{fig:rho}

\end{figure}

\textbf{Ablation study on the value of $\rho$.}
We employ $\rho$ to control the intervals at which ISG performs sampling. When the value of $\rho$ is too small, meaningful styles may be excluded, resulting in sub-optimal generalization capability. In contrast, meaningless styles can be introduced when the value of $\rho$ is too large. As illustrated in Fig. ~\ref{fig:rho}, the optimal value of $\rho$ is 3.

\begin{table}[tp]
\centering
\caption{Ablation study on the sampling strategy in ISG. Here, Gaussian sampling and uniform sampling are compared.}
\label{tab:GBSvsUBS}
\vspace{-1em}
\begin{tabular}{c||cc|cc}
\toprule[1pt]
Sampling & \multicolumn{2}{c|}{D+C3+MS$\to$M} & \multicolumn{2}{c}{M+D+C3$\to$MS} \\ \cline{2-5} 
Strategy & mAP & Rank-1 & mAP & Rank-1 \\ \hline \hline
Baseline  &59.3  &81.2  &  14.7 &  35.2 \\\hline
Gaussian & 63.9    &84.8        &19.5     &44.2        \\
Uniform  & 65.8    &86.2        &20.2     &45.7        \\ \toprule[1pt]
\end{tabular}
\vspace{-1em}
\end{table}

\begin{figure}[tp]
\setlength{\abovecaptionskip}{0pt}
\centering
\includegraphics[width=\linewidth]{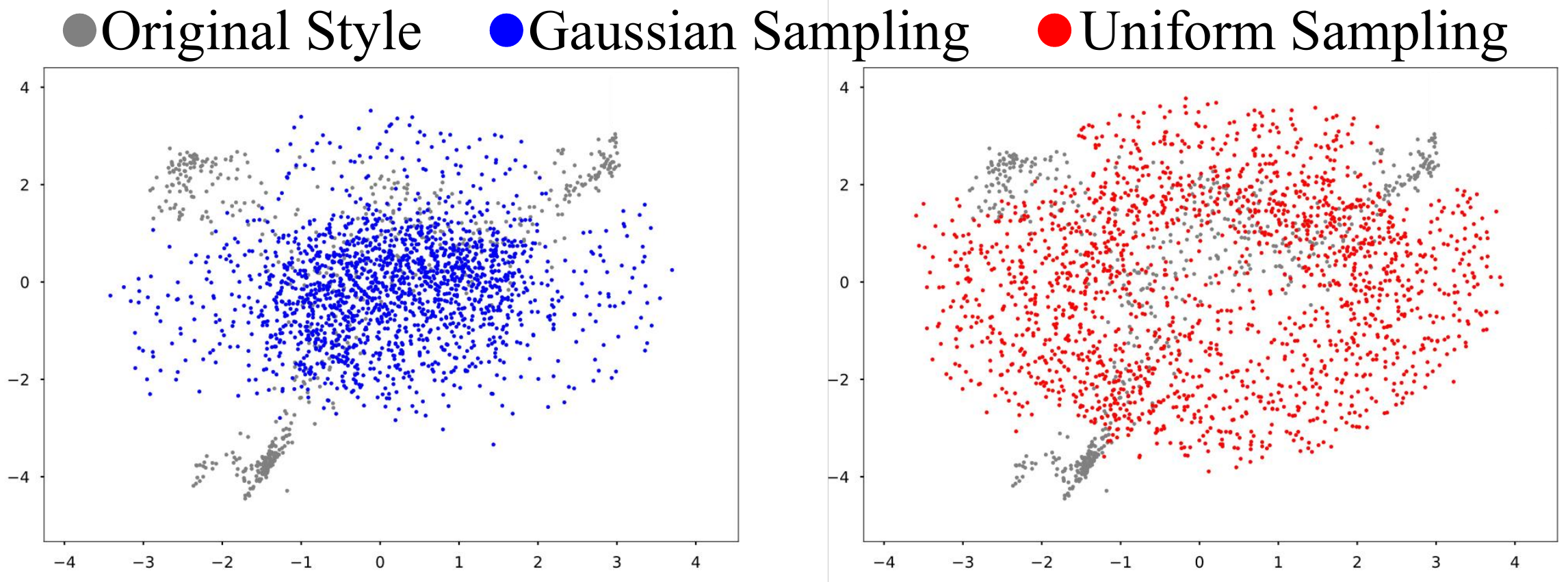}
\caption{The 2-D t-SNE \cite{van2008visualizing} visualization of feature styles. We concatenate the channel-wise mean and standard deviation of a feature map to represent its style. Best viewed in color.}
\label{fig:GvsU_tsne}

\end{figure}

\begin{table}[tp]
\centering
\caption{
Ablation study on the order of forward and backward propagations. ``F'' and ``B'' stand for the forward and backward propagations, respectively. }
\label{tab:forward}
\vspace{-1em}
\begin{tabular}{c||cc|cc}
\toprule[1pt]
\multirow{2}{*}{Variant} & \multicolumn{2}{c|}{D+C3+MS$\to$M} & \multicolumn{2}{c}{M+D+C3$\to$MS} \\ \cline{2-5} 
 & mAP & Rank-1 & mAP & Rank-1 \\ \hline \hline
Baseline  &59.3  &81.2  & 14.7 &  35.2 \\\hline
FFB &64.7  &85.6  &19.9  &44.4  \\
FBF &65.8  &86.2  &20.2  &45.7  \\ \toprule[1pt]
\end{tabular}

\end{table}

\begin{table}[tp]
\centering
\caption{Ablation study on the position in which ISG is applied.}
\label{tab:stage}
\vspace{-1em}
\begin{tabular}{c||cc|cc}
% \hline
\toprule[1pt]

\multicolumn{1}{c||}{\multirow{2}{*}{Stage}} & \multicolumn{2}{c|}{D+C3+MS$\to$M} &\multicolumn{2}{c}{M+D+C3$\to$MS}\\ \cline{2-5} 
 & mAP & Rank-1 & mAP & Rank-1 \\ \hline \hline
baseline & 59.3 & 81.2 &  14.7 &  35.2\\ \hline
after stage1 & 65.8 & 86.2 &20.2&45.7\\
after stage12 & 63.8 & 84.5 &19.9&45.4\\
after stage123 & 63.3 & 84.7 &20.0&44.7\\
after stage1234 & 8.9 & 23.5 &1.8&6.7\\ \toprule[1pt]
\end{tabular}
\end{table}

\begin{table}[htp]
\centering
\caption{
Comparisons of model complexity. $\ddag$ indicates that the meta-learning scheme is adopted during training.}
\label{tab:complexity}
\vspace{-1em}
\resizebox{0.45\textwidth}{!}{%
\begin{tabular}{c||ccc}
\toprule[1pt]
Method & Train & Inference & Params \\ \hline
\hline
% RaMoE$^\ddag$ \cite{dai2021generalizable}&0.989s/iter  &0.94ms/img  &39.3M   \\
M$^3$L$^\ddag$ \cite{zhao2021learning} &1.974s/iter  &0.21ms/img  &23.5M   \\
MetaBIN$^\ddag$ \cite{choi2021meta}&0.585s/iter  & 0.32ms/img  &23.6M    \\ 
ACL$^\ddag$ \cite{zhang2022acl}&0.588s/iter  &0.36ms/img  &29.6M    \\ \hline
% META  \cite{xu2022meta}&0.603s/iter  &0.80ms/img  &44.3M    \\ \hline
Baseline &0.295s/iter  &0.21ms/img  &23.5M    \\
IL w/o ISG &0.347s/iter  &0.21ms/img  &23.5M    \\
IL &0.352s/iter  &0.21ms/img  &23.5M   \\ \toprule[1pt]
\end{tabular}}
\end{table}

\textbf{Ablation study on the sampling strategy}. 
We test the performance of two different sampling strategies for ISG, namely Gaussian sampling and uniform sampling. The results are presented in Table \ref{tab:GBSvsUBS}.
As can be seen from the results, uniform sampling is superior to Gaussian sampling.  This is because styles that are sampled with equal probability can be more diverse and are less likely to be correlated with the original styles in the training images. Therefore, uniform sampling is more suitable for application in an IL context.

\begin{figure*}[htp]
\setlength{\abovecaptionskip}{0pt}
\centering
\includegraphics[width=0.975\linewidth]{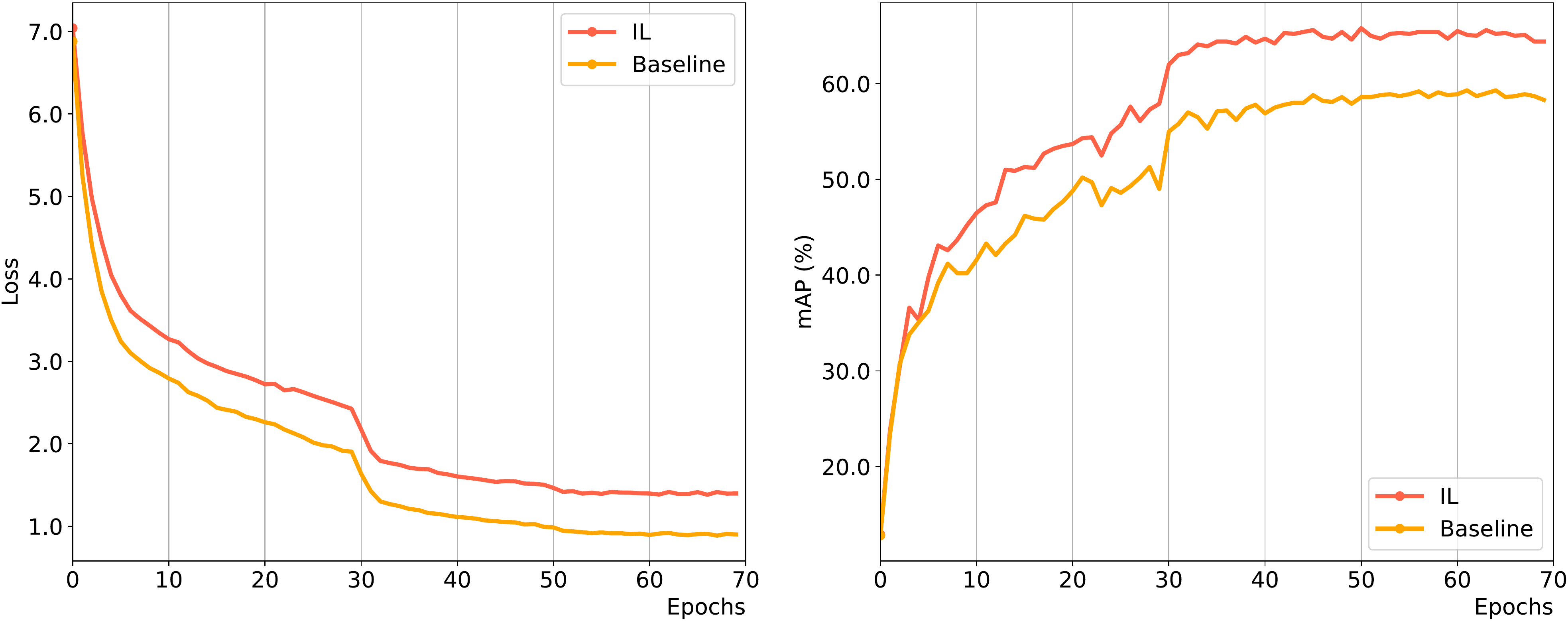}
\caption{The average training loss (left) and inference mAP curves (right). Experiments are conducted under the D+C3+MS$\to$M setting. Best viewed in color. 
 }
\label{fig:loss_map}
\end{figure*}

\textbf{The order of forward and backward propagations}.
A variant of IL involves the second forward propagation being moved to the position between the first forward propagation and the backward propagation. 
We compare the performance of these two schemes in Table \ref{tab:forward} and find that our proposed scheme achieves better performance. This may be because the updated feature extractor $f_{{\theta}^{'}}$ produces more discriminative features after the backward propagation has been performed, which improves the quality of the prototypes stored in the memory banks. 

\textbf{The position in which ISG is applied.} We place ISG in different stages of the ResNet50 model and compare their performance in Table \ref{tab:stage}. As the table shows, the best performance is achieved when ISG is placed after the first stage of ResNet50. Placing ISG behind more stages does not further improve performance. This may be because, when the hyper-parameter $\rho$ is properly set, it is possible to produce sufficiently diverse styles using only one ISG module. When the ISG module is placed after stage4, the performance degrades dramatically; this is because the features produced by stage4 contain rich semantic information. In comparison, the bottom CNN layers (\textit{e.g.}, layers in stage 1) contain more style information, as is also verified in \cite{jin2020style,zhou2021domain}.

\textbf{Comparisons of model complexity.}\label{complex}
 In this experiment, we demonstrate that IL not only achieves superior performance in terms of ReID accuracy, but is also advantageous in terms of time and space complexities. To facilitate a fair comparison, we utilize the same batch size and the same TITAN Xp GPU for all methods in Table \ref{tab:complexity}. The results show that the computational cost of interleaved learning is significantly lower than that of meta-learning-based methods in the training stage. This is because meta-learning requires two backward propagations, resulting in a high computational cost. Moreover, the time cost introduced by the second forward pass in each iteration is found to be very small. It is also worth noting that the ISG module itself introduces a near-negligible cost of 0.005s/iter. During testing, we remove the ISG module from the feature extractor; therefore, it is used as a single standard backbone model, and the test speed is very fast.

\subsection{Qualitative Analysis}

\textbf{Comparisons in training loss and inference mAP curves.} 
This experiment compares the average training loss and inference mAP curves between the IL framework and the baseline. The comparisons are illustrated in Fig. \ref{fig:loss_map}. It can be observed that the training loss of the IL framework decreases slower than that of the baseline. However, the inference mAPs of the IL framework are substantially higher. This is because the feature extractor and classifiers are updated with the features of interleaved styles in the IL framework, which reduces the model's risk of overfitting on the source domain data and promotes the model's ability to generalize to the target domain data.

\begin{figure*}[htp]
\setlength{\abovecaptionskip}{0pt}
\centering
\includegraphics[width=0.975\linewidth]{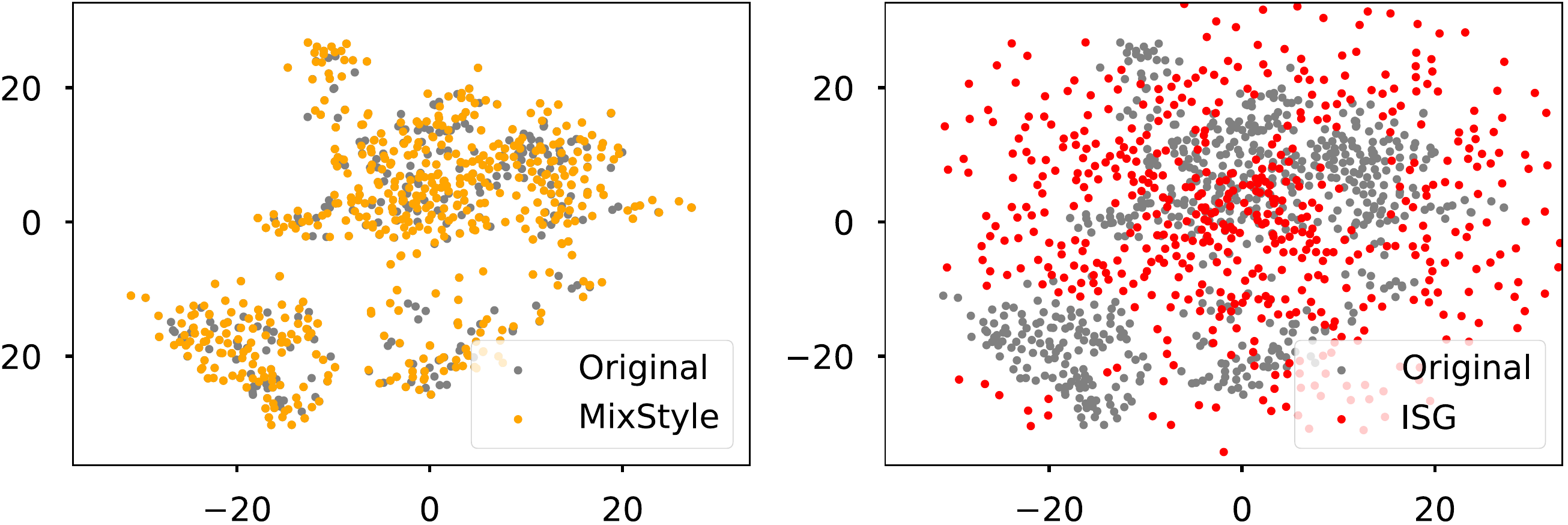}
\caption{T-SNE \cite{van2008visualizing} visualization of the styles produced by MixStyle \cite{zhou2021domain} and ISG. `Original' represents the original styles of the source images in one batch. Best viewed in color. 
 }
\label{fig:MixStyle}
\end{figure*}

\begin{figure*}[htp]
\setlength{\abovecaptionskip}{0pt}
\centering
\includegraphics[width=0.975\linewidth]{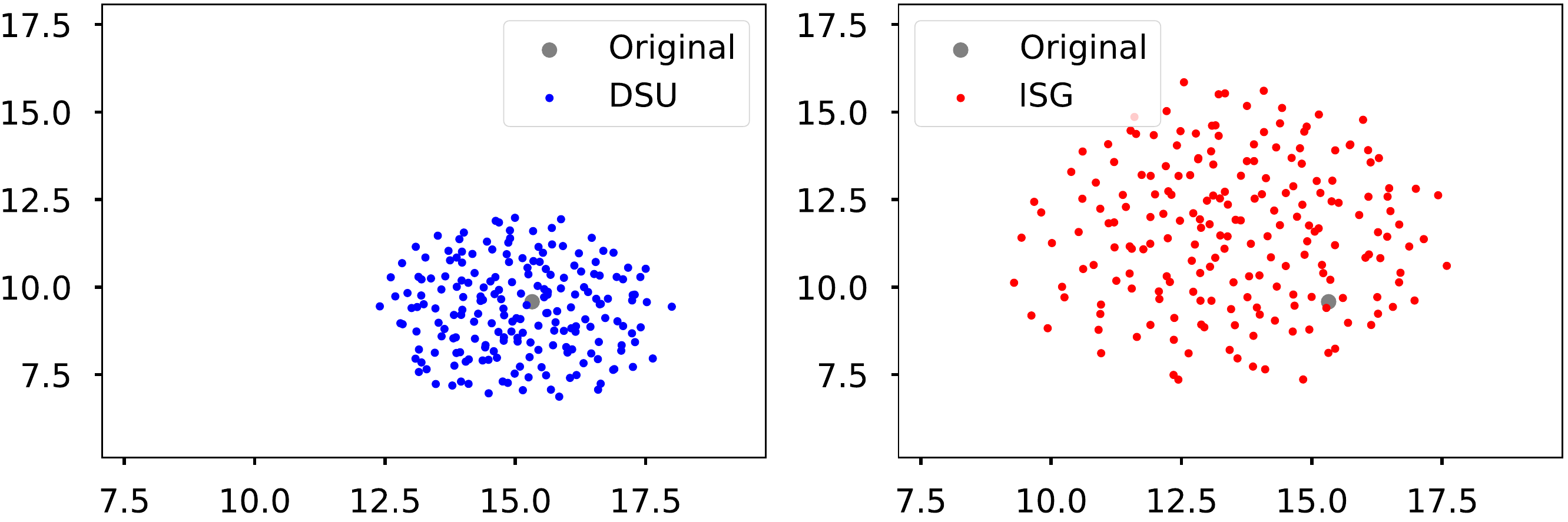}
\caption{T-SNE \cite{van2008visualizing} visualization of the styles produced by the DSU \cite{li2022uncertainty} and ISG. We randomly select a sample from the Market-1501 database and denote its style as `Original' in the figure. To illustrate their essential difference, we repeatedly perform DSU and ISG for this original style for 100 times, respectively. Best viewed in color. 
}
\label{fig:DSU}
\end{figure*}

\textbf{Comparisons between ISG and MixStyle.}
We visualize the styles produced by ISG and compare them with those generated by MixStyle \cite{zhou2021domain} in Fig. \ref{fig:MixStyle}. As the figure shows, ISG creates more diverse feature styles. More specifically, the styles generated by MixStyle are quite similar to the original styles; in comparison, the styles generated by ISG are scattered uniformly within the estimated  intervals (Eq.~\ref{equ:sample_new_style1},\ref{equ:sample_new_style2}) and are therefore more diverse.

\textbf{Comparisons between ISG and DSU.}
We visualize the styles generated by ISG and DSU \cite{li2022uncertainty} in Fig.~\ref{fig:DSU}. It can be observed that the styles generated by DSU are correlated with the original style, while ISG creates styles that are more independent from the original style. The main reason is that DSU conducts Gaussian sampling centered on each original sample, while ISG conducts uniform sampling within much wider intervals centered on the mean style of all samples in the batch.

MixStyle and DSU are successful feature stylization methods for data augmentation purposes. However, their synthesized feature styles lack independence from the original styles. In comparison, our proposed ISG method exhibits good compatibility with the IL framework, where independence plays a vital role. Indeed, the experimental results in Table \ref{tab:AUGvsIL} clearly show that ISG outperforms these two methods by significant margins.

\section{Conclusion} \label{conclusion}

In this paper, we propose a novel style interleaved learning framework for domain generalizable person ReID (DG ReID). This learning strategy adopts the features of different styles for classifier updating and loss computation, which prevents the feature extractor from overfitting to the existing feature styles contained in the source domains. We further introduce a new feature stylization approach, ISG, which can produce more diverse styles that are independent from the original styles in a batch. Extensive experiments demonstrate that our approach consistently outperforms the state-of-the-art methods by significant margins. Although IL and ISG are simple and efficient techniques, they still have some limitations: (1) The diversity of styles generated by ISG can vary depending on the source styles and the hyperparameter $\rho$. (2) Despite IL achieving state-of-the-art performance, it falls short in terms of generalization on specific datasets, such as the extensive MSMT17 dataset, when compared to supervised approaches. This observation motivates us to develop more powerful DG ReID methods in the future.

\section{Acknowledgments}

This work was supported by the National Natural Science Foundation of China under Grant 62076101, Guangdong Basic and Applied Basic Research Foundation under Grant 2023A1515010007, the Guangdong Provincial Key Laboratory of Human Digital Twin under Grant 2022B1212010004, and the Program for Guangdong Introducing Innovative and Entrepreneurial Teams under Grant 2017ZT07X183.

\section{Appendix}
This appendix provides more details about the feature stylization methods compared in this paper.

\emph{pAdaIN \cite{nuriel2021permuted}.}
This method swaps the styles between two samples in the batch. Specifically,
\begin{equation} \label{eq:pAdaIN}
\operatorname{pAdaIN}(\bm{F}) = \sigma(\bm{F}') \odot \frac{\bm{F} - \mu(\bm{F})}{\sigma(\bm{F})} + \mu(\bm{F}'),
\end{equation}
where $\bm{F}$ and $\bm{F}'$ are feature maps of two different samples in a batch. 

pAdaIN only employs styles present in the source data; therefore, the styles it employs lack diversity compared with those created by ISG.

\emph{MixStyle \cite{zhou2021domain}.}
This method mixes the statistics of two feature maps in a linear manner. Specifically,
\begin{align}
\bm{\beta}_{MS} &= \lambda \mu(\bm{F}) + (1 - \lambda) \mu(\bm{F}'),
 \label{eq:mix_gamma} \\
\bm{\gamma}_{MS} &= \lambda \sigma(\bm{F}) + (1 - \lambda) \sigma(\bm{F}'), \label{eq:mix_beta}
\end{align}
where $\lambda$ is a weight that is randomly sampled from the beta distribution. Finally, $\bm{\gamma}_{MS}$ and $\bm{\beta}_{MS}$ are applied to $\bm{F}$ to change its style,
\begin{equation} \label{eq:mixstyle}
\operatorname{MixStyle}(\bm{F}) = \bm{\gamma}_{MS} \odot \frac{\bm{F} - \mu(\bm{F})}{\sigma(\bm{F})} + \bm{\beta}_{MS}.
\end{equation}

Although MixStyle generates new feature styles that do not exist in the source data, the newly synthesized feature styles are closely correlated with the original styles.
% In comparison, our ISG method presents more diversity compared with MixStyle.

\emph{DSU \cite{li2022uncertainty}.}
This approach adds perturbations to the original style of one sample to facilitate the synthesis of new styles. The perturbations are generated according to the following method:
\begin{align}
\bm{\beta}_{DSU} &= \mu(\bm{F}) + \epsilon_{\mu}\hat{\bm{\sigma}}_{\mu},
 \label{eq:mix_gamma} \\
\bm{\gamma}_{DSU} &= \sigma(\bm{F}) + \epsilon_{\sigma}\hat{\bm{\sigma}}_{\sigma}, \label{eq:mix_beta}
\end{align}
\begin{equation} \label{eq:mixstyle}
\operatorname{DSU}(\bm{F}) = \bm{\gamma}_{DSU} \odot \frac{\bm{F} - \mu(\bm{F})}{\sigma(\bm{F})} + \bm{\beta}_{DSU},
\end{equation}
where $\epsilon_{\sigma},\epsilon_{\mu} \sim \mathcal{N}(\textbf{0},\textbf{1})$.

The new styles created by DSU are still correlated with the original style. This is because DSU performs Gaussian sampling centered on one original feature style.

% \section{Acknowledgments}

% This work was supported by the National Natural Science Foundation of China under Grant 62076101, Guangdong Basic and Applied Basic Research Foundation under Grant 2023A1515010007, the Guangdong Provincial Key Laboratory of Human Digital Twin under Grant 2022B1212010004, and the Program for Guangdong Introducing Innovative and Entrepreneurial Teams under Grant 2017ZT07X183.

\bibliographystyle{IEEEtran}
\bibliography{egbib}

\end{document}